%% file: main.tex
\pdfoutput=1
\documentclass[11pt]{article}
\usepackage{authblk}

\usepackage{emnlp2023}

\usepackage{times}
\usepackage{latexsym}

\usepackage[T1]{fontenc}
\usepackage[utf8]{inputenc}

\usepackage{microtype}

\usepackage{booktabs}

\usepackage{multirow}
\usepackage{pifont}
\usepackage[justification=centering]{caption}
\usepackage{enumitem}
\usepackage{array,multirow,graphicx}
\usepackage{float}
\usepackage{amsmath}

\usepackage{subcaption}

\usepackage[symbol]{footmisc}

\usepackage{placeins}

\newcommand{\hidetodo}[1]{{\color{red}{\small\bf\sf}}}

\newcommand{\hidelogan}[1]{{\color{olive}{\small\bf\sf}}}

\newcommand\Tstrut{\rule{0pt}{2.6ex}}         % = `top' strut
\newcommand\Bstrut{\rule[-1.3ex]{0pt}{0pt}}   % = `bottom' strut

\title{Efficient Transformer Knowledge Distillation: A Performance Review}

\usepackage{etoolbox}
\makeatletter
\patchcmd{\maketitle}
 {\def\@makefnmark}
 {\def\@makefnmark{}\def\useless@macro}
 {}{}
\makeatother

\author[1]{Nathan Brown}
\author[1]{Ashton Williamson}
\author[1]{Tahj Anderson}
\author[2,*]{Logan Lawrence\thanks{$^*$Corresponding author.}}
\affil[ ]{$^1$School of Computing, Clemson University, $^2$Giant Oak Inc.}
\affil[ ]{\texttt{\{nbrown9,taw2,tahja\}@clemson.edu,logan.lawrence@giantoak.com}}

\begin{document}
\maketitle

\input{sections/0_abstract}
\input{sections/1_introduction}
\input{sections/2_related_work}
\input{sections/3_method}
\input{sections/4_experiments}
\input{sections/5_conclusion}
\input{sections/7_acknowledgements}
\input{sections/8_limitations}

\bibliographystyle{acl_natbib}
\bibliography{references}

\appendix

\input{sections/6_appendix}

\end{document}

%% file: sections/0_abstract.tex
\begin{abstract}
As pretrained transformer language models continue to achieve state-of-the-art performance, the Natural Language Processing community has pushed for advances in model compression and efficient attention mechanisms to address high computational requirements and limited input sequence length. Despite these separate efforts, no investigation has been done into the intersection of these two fields. In this work, we provide an evaluation of model compression via knowledge distillation on efficient attention transformers. We provide cost-performance trade-offs for the compression of state-of-the-art efficient attention architectures and the gains made in performance in comparison to their full attention counterparts. Furthermore, we introduce a new long-context Named Entity Recognition dataset, GONERD, to train and test the performance of NER models on long sequences. We find that distilled efficient attention transformers can preserve a significant amount of original model performance, preserving up to \textbf{98.6\%} across short-context tasks (GLUE, SQUAD, CoNLL-2003), up to \textbf{94.6\%} across long-context Question-and-Answering tasks (HotpotQA, TriviaQA), and up to \textbf{98.8\%} on long-context Named Entity Recognition (GONERD), while decreasing inference times by up to \textbf{57.8\%}. We find that, for most models on most tasks, performing knowledge distillation is an effective method to yield high-performing efficient attention models with low costs. 
\end{abstract}

%% file: sections/1_introduction.tex
\section{Introduction}
The rise of Transformer-based models \cite{vaswani2017attention} has driven significant advancements in the field of Natural Language Processing (NLP). Of these models, BERT \cite{devlin2018bert, bertology} produced landmark performance in a variety of NLP tasks such as Question Answering (QA), Named Entity Recognition (NER), and GLUE \cite{wang2019glue}. BERT-based models \cite{bertology} continue to dominate the field \cite{zhou2023comprehensive} with variations such as RoBERTa \cite{liu2019roberta} dramatically improving performance on downstream tasks.
 
However, BERT-based models often have a fairly short maximum input length of 512 tokens, severely limiting their capabilities in long-context situations. Attempting to increase this limit to allow for longer sequences often results in significantly greater computational requirements. This has given rise to the creation of \emph{efficient} attention transformer models \cite{tay2022efficient} such as Longformer \cite{Beltagy2020Longformer}, Big Bird \cite{zaheer2020bigbird}, Nyströmformer \cite{nystromformer}, and LSG \cite{condevaux2022lsg}, which can accept as input much longer sequences with reduced computational overhead by modifying and approximating BERT's original attention mechanism.

While efficient attention models require less computational resources on long-context tasks when compared to their non-efficient counterparts, they are still often computationally expensive to train and deploy \cite{sharir2020cost}. Thus, organizations and individuals are required to grapple with increased operational costs, difficulty deploying these models on resource-limited hardware such as mobile devices, and often must rely on cloud-based solutions which restricts model availability in scenarios with limited internet access. 

In response to computational challenges associated with transformer models, the NLP community has invested considerable efforts into creating cheaper yet performant models. This has been particularly the case in the study of Knowledge Distillation (KD) \cite{gou2021knowledge, hinton2015distilling}. However, despite the rapid progress of KD and its effectiveness in model compression, little work has been done toward the investigation of the intersection of KD and efficient attention architectures. As such, we focus on combining these two methodologies. We believe this is an essential effort for creating models that can cheaply and effectively operate on a production scale on long-context tasks. Furthermore, despite its significance in practical NLP usage, Named Entity Recognition (NER) still does not have a well-accepted long-context benchmark. Our work attempts to address these two needs directly.

The main contributions of this work are twofold: 
\begin{enumerate}
    \item Performance analysis of popular pretrained efficient transformers and their distilled students in various contexts, including \textbf{GLUE}, \textbf{SQuAD} \cite{squad, squadv2}, \textbf{HotpotQA} \cite{yang2018hotpotqa}, \textbf{TriviaQA} \cite{triviaqa}, \textbf{CoNLL-2003} \cite{conll-2003}, and \textbf{GONERD}.
    \item The introduction of a new long-context NER task: the \textbf{G}iant \textbf{O}ak \textbf{NER} \textbf{D}ataset (\textbf{GONERD}).
    This dataset and all models are publicly available on Hugging Face\footnote{\url{https://huggingface.co/giant-oak}}.
\end{enumerate}

In particular, we find that distilling Longformer-RoBERTa \cite{Beltagy2020Longformer} yields the best results during our experiments, producing substantial improvements in cost performance over state-of-the-art models. In short, it retains considerable performance on GLUE (\textbf{92.3\%}), SQuAD (\textbf{93.0\%}), HotpotQA (\textbf{88.4\%}), CoNLL-2003 (\textbf{99.8\%}), and GONERD (\textbf{95.9\%}) while decreasing inference times by \textbf{49.3\%} on long sequences. In the context of GONERD, this is effectively \textbf{95.9\%} of the original model's performance for \textbf{50.7\%} of the cost. 

%% file: sections/2_related_work.tex
\section{Related Work}

Considerable success has been made in the compression of BERT \cite{devlin2018bert} which, at the time of its release, was one of the largest models in NLP. BERT itself has been expanded to fit many different use cases including, but not limited to, RoBERTA \cite{liu2019roberta}, a model built to improve BERT performance on a variety of tasks through clever choices in training data and hyperparameters, XLM-R \cite{xlmr}, which was built using similar methods on extremely multilingual data (100 languages), and DistilBERT \cite{sanh2019distilbert}, which sought to greatly reduce the computational costs of BERT through knowledge distillation. ALBERT \cite{lan2019albert} factorizes the embedding matrices of BERT and shares weights between layers to significantly decrease the parameter size, thereby decreasing training and inference costs.

BERT-based distillation methods, such as DistilBERT \cite{sanh2019distilbert}, TinyBERT \cite{jiao2020tinybert}, and MobileBERT \cite{sun2020mobilebert} have gained prominence due to their utilization of distillation techniques and can be applied to a wide variety of BERT-based architectures. These models have significantly reduced the computational requirements and resource consumption associated with BERT-based NLP models, making them more accessible and easily deployable on resource-constrained hardware. However, BERT's attention mechanism still results in a quadratic dependency on sequence length, resulting in greater computational requirements at higher sequence lengths.

To solve this problem with BERT-based architectures, methods have been developed to create \emph{efficient} attention transformer models \cite{tay2022efficient} which can operate on sequences many times longer than their BERT counterparts. Two popular methods in this area are Longformer \cite{Beltagy2020Longformer} and Big Bird \cite{zaheer2020bigbird}, which use dilated sliding window and a combination of global, sliding, and random activations in their attention matrices, respectively, to increase the maximum input sequence length from 512 to 4096. More recently, Local Sparse Global (LSG) \cite{condevaux2022lsg} attention uses a Locality Sensitive Hashing algorithm \cite{practical_lsh} with the Local, Sparse, and Global patterns used in Longformer and Big Bird, whereas Nystr\"omformer \cite{nystromformer} uses a Nystr\"om matrix approximation to the regular softmax attention, reducing self-attention complexity to linear time. 

The \textbf{L}ong-\textbf{R}ange \textbf{A}rena (\textbf{LRA}) \cite{tay2020long}, a comprehensive suite of benchmarking tasks toward systematically evaluating long-context transformer architectures, is novel in that its tasks largely decouple the effect of Masked Language Modeling (MLM) pretraining from efficient model performance. While useful for developing new transformer architectures, we are primarily focused on the comparative performance between student and teacher models on downstream tasks after having been pretrained/distilled on MLM. As such, LRA is not utilized in this paper.

%% file: sections/3_method.tex
\section{Methodology}
\subsection{Knowledge Distillation}
\label{subsection:kd}

Knowledge Distillation (KD) for transformer-based architectures \cite{gou2021knowledge} is most commonly executed in three steps: (1) Pretrain a larger, complex model. (2) Distill knowledge from the larger complex model into a smaller, simpler model. (3) Fine-tune the student model on a downstream task. While effective in short-context scenarios, this three-step process leaves room for ambiguity regarding the recommended distillation process for long-context efficient attention transformer models.

In this paper, we use the term \textit{"convert"} to refer to the process of updating a pretrained LM to use an efficient attention pattern, i.e. one capable of input lengths longer than 512 tokens. Considering this, we can identify two possible methods of inserting the conversion operation into the classical KD pipeline:
\begin{enumerate}
    \item \textbf{Convert-Then-Distill}: Convert teacher $\rightarrow$ Pretrain teacher $\rightarrow$ Distill into student $\rightarrow$ Fine-tune student on downstream task
    \item \textbf{Distill-Then-Convert}: Pretrain teacher $\rightarrow$ Distill into student $\rightarrow$ Convert student $\rightarrow$ Fine-tune student on downstream task
\end{enumerate}

Although \textbf{Distill-Then-Convert} is conceptually interesting and potentially fruitful, we will be only covering \textbf{Convert-Then-Distill} in this work. However, we do include an experiment directly extending the maximum input sequence length in Section \ref{subsection:extending_input_sequence_length} of existing non-efficient distilled students, demonstrating the necessity of the \textit{conversion} step in long-context tasks in terms of reducing a model's computational requirements.

Within the realm of \textbf{Convert-Then-Distill}, we perform knowledge distillation using the same process utilized in the creation of DistilBERT \cite{sanh2019distilbert}. Namely, we begin by compressing pretrained efficient teacher models Longformer RoBERTa \cite{Beltagy2020Longformer}, Big Bird RoBERTa \cite{zaheer2020bigbird}, LSG RoBERTa \cite{condevaux2022lsg}, and Nyströmformer \cite{nystromformer}. In all cases, the number of hidden layers is reduced by a factor of two, with the student model being initialized by taking every other hidden layer from the teacher.

During training, the distillation loss is calculated over the soft target probabilities of the teacher. A softmax temperature is used, and a linear combination of the distillation loss, MLM supervised training loss, and cosine embedding loss is performed. For additional details, see Appendix \ref{appendix:kd_details}.

\subsection{Distillation Datasets}
To perform knowledge distillation, we utilize the Open Super-large Crawled Aggregated coRpus project (OSCAR) \cite{OrtizSuarezSagotRomary2019}, a large open-source corpus of raw unannotated web text. MLM pretraining, and by consequence Knowledge Distillation, requires a large amount of text data \cite{pretraining_survey} and OSCAR allows for the selection of a large amount of high-quality long-context text samples. This dataset is used during distillation alongside the commonly used training dataset, BookCorpus \cite{bookcorpus}. The selection of these two distillation datasets was determined through an experiment investigating the effect of different distillation datasets on downstream performance, as seen in Section \ref{subsection:effect_of_distillation_data}.

When constructing our data to be used for knowledge distillation, we first filtered out all data from the OSCAR23.01 corpus which was not classified as having an eighty percent or higher chance of being English text to align with downstream tasks. To seek out only high-quality data, we also remove samples with quality annotations indicating tiny, short, or noisy sequences. We remove any data with a harmful perplexity score of 13.51 or less \cite{2212.10440} using perplexity scores provided by the OSCAR corpus \cite{OrtizSuarezSagotRomary2019}, and additionally remove any harmful categories. Finally, we select a sample from our filtered dataset to be used during distillation consisting of nearly three million sequences, then distil using this OSCAR subset alongside BookCorpus \cite{bookcorpus}, totaling 19 GB of uncompressed text.

\subsection{GONERD}
\input{tables/length_stats_conll_gonerd} 
Data for GONERD (\textbf{G}iant \textbf{O}ak \textbf{NER} \textbf{D}ataset) was obtained by web scraping articles from publicly available sources such as online news and press release websites prior to being sampled and hand-labeled. A combination of automatic and manual filtering was then applied to remove text containing code and other unwanted data such as sequences deemed short, noisy, or duplicates.

As the explicit goal of GONERD is to gauge the performance of long-context NER models, we briefly quantify what sequence lengths are present within the dataset. We compare against CoNLL-2003, shown in Table \ref{table:length_stats_conll_gonerd}, as it is widely used throughout NER literature. We find, on average, GONERD has much longer sequences than CoNLL-2003 (\textbf{507.6} vs. \textbf{14.5}), with a right skew as seen by the 50\% percentile (\textbf{330}) being lower than the mean. We show this skew in Appendix \ref{appendix:gonerd_eda}. 

Furthermore, we find that approximately \textbf{35\%} of GONERD sequences are above the 512 token threshold, whereas none of the CoNLL-2003 sequences occur in this range. Finally, we find that \textbf{0.2\%} of sequences are longer than 4096, which are truncated at training and inference time. For more information on GONERD, including exploratory data analysis and additional comparisons with CoNLL-2003, see Appendix \ref{appendix:gonerd_data_collection} and \ref{appendix:gonerd_eda}. 

\subsection{LSG RoBERTa Pretraining}

Although the implementation of LSG RoBERTa \cite{condevaux2022lsg} is publicly available, there are currently no publicly accessible weights, neither compressed nor uncompressed, that have been pretrained on long-context sequences. While analysis on inference and memory utilization can be performed without these weights, undergoing a comprehensive performance analysis of LSG RoBERTa or utilizing this model in research or production requires further pretraining.

To address this issue, and to yield a pretrained teacher model as the first step towards developing a distilled student model, we pretrain a randomly initialized LSG RoBERTa model using the same dataset presented in LSG's inception \cite{condevaux2022lsg}. This consists of English Wikipedia, BookCorpus, and CC\_News.

%% file: tables/length_stats_conll_gonerd.tex
\begin{table}[ht!]
\setlength{\belowcaptionskip}{0pt}
\setlength{\tabcolsep}{8pt}
\begin{center}
{\footnotesize
\begin{tabular}{lcc}
\hline \Tstrut
Length & \textbf{CoNLL-2003} (512) & \textbf{GONERD} (4096) \\[.4ex]
\hline \Tstrut
mean & 14.5 & 507.6 \\
std. dev. & 11.8 & 556.5 \\
min & 1.0 & 1.0 \\
25\% & 6.0 & 170.0 \\
50\% & 10.0 & 330.0 \\
75\% & 22.0 & 658.0 \\ \Bstrut
max & 124.0 & 6768.0 \\ 
\hline
\end{tabular}
\vspace{.3cm}
\caption{Summary statistics on sequence length of CoNLL-2003 and GONERD. All statistics are computed over the whole dataset. "mean" and "std. dev." follow their usual definitions, "min" and "max" are the lengths of the shortest and longest datasets. "25\%", "50\%", "75\%" are the 25th, 50th, and 75th percentiles, respectively.}
\label{table:length_stats_conll_gonerd}
}
\end{center}
\end{table}

%% file: sections/4_experiments.tex
\section{Experiments}
\subsection{Inference Speed and Memory Usage}
\label{subsection:inference_speed_text}

We calculate the average inference time and maximum GPU memory utilization for a variety of short-context and long-context transformer models as a proxy for predicting costs for hosting each model type in production, as displayed in Table \ref{table:inference_memory}. Moreover, we compare the potential cost of deploying efficient attention models versus their distilled equivalents. All models were tested in a uniform environment utilizing a single 80GB A100 GPU with a sequence length of either 512 or 4096 tokens and a batch size of 16. 
\input{tables/inference_memory}
\input{tables/glue_performance}

We find an average \textbf{45.2\%} decrease in inference times for long-context efficient attention models and an average \textbf{2.6\%} percent decrease in GPU memory utilization across all distilled efficient models. Among the distilled efficient students, Longformer produces the fastest inference speed and least peak GPU memory usage in both 512 and 4096 settings, despite having the most parameters.

We find that KD as discussed in Section \ref{subsection:kd} does not significantly impact peak GPU memory usage during inference across both efficient (LSG, Nystr\"omformer, Longformer, Big Bird) and non-efficient (DistilBERT, DistilRoBERTa) architectures. Larger modifications to the student architecture (TinyBERT, MobileBERT, ALBERT), produce varying speeds and levels of GPU memory usage.

\subsection{Extending Input Sequence Length}
\label{subsection:extending_input_sequence_length}

To demonstrate the necessity of efficient attention architectures, we investigate the feasibility of using existing models on long-context tasks by allowing inefficient attention models to process longer sequence lengths. To explore this, as presented in Table \ref{table:ablation_extending}, we examine the inference speed and peak GPU memory consumption during inference on full attention BERT-based models after being adjusted to compute sequence lengths of up to 4096 tokens, employing the same benchmarking methodology as seen in Section \ref{subsection:inference_speed_text}.

\input{tables/extend_input_performance}

Our findings illustrate a noticeable trend: although it is possible to allow inefficient models to accept input sequences of up to 4096 tokens, there are significant speed and memory costs associated with doing so. Moreover, the newly initialized position embeddings would require anyone using these extended models to perform additional pretraining to yield acceptable long-context performance - a process that would be slower and more expensive than training an efficient attention model. This difficulty training would also inherently transfer to the process of fine-tuning these models on downstream tasks.

This evidence suggests that, although it is possible for full attention models to operate in long-context scenarios, it is often associated with increased inference and training costs when compared to non-distilled and distilled efficient attention models. As such, efficient attention models are an important step toward reducing the operational costs of long-context models, and distillation after conversion can be a useful methodology to further reduce costs and improve model accessibility.

\subsection{Performance Benchmarks}
\label{subsection:performance_benchmarks}

\paragraph{GLUE} We perform hyperparameter optimization using Population-Based Training \cite{jaderberg2017population} on several baselines, augmented, efficient attention, and distilled efficient attention models on the GLUE benchmark \cite{wang2019glue}. As seen in Table \ref{table:glue_results}, we find that distilling efficient attention models yields compressed models capable of retaining, on average, \textbf{94.6\%} of teacher model performance across all GLUE tasks and metrics. Distilled Big Bird produces the highest GLUE scores on average when compared to our distilled efficient attention models. Distilled Nystromformer sees a slight increase in performance when compared to its teacher. Distilled LSG retains only \textbf{87.3\%} of teacher performance.

\input{tables/qa_performance}
\paragraph{Question Answering} We train and evaluate all short-context and long-context transformer models on SQuAD1.1 \cite{squad}. Moreover, we train and evaluate all long-context transformer models using a maximum sequence length of 4096 tokens on TriviaQA \cite{triviaqa} and HotpotQA \cite{yang2018hotpotqa} for up to 5 and 10 epochs, respectively. Results are reported in Table \ref{table:qa_performance}.

We find that on SQuAD, HotpotQA, and TriviaQA, efficient attention students retained up to \textbf{94.8\%}, \textbf{94.1\%}, and \textbf{95.0\%} of original model F1 performance, respectively. LSG RoBERTa was particularly strongly affected by the distillation process on long-context Question and Answering tasks, preserving \textbf{75.7\%} of teacher performance on HotpotQA and \textbf{57.9\%} on TriviaQA. Distilled Nystr\"omformer retains the most performance from its teacher with an average of \textbf{94.7\%} across all QA benchmarks, but it is still outperformed by distilled Longformer by \textbf{2.3\%}.

\input{tables/ner_performance}
\paragraph{Named Entity Recognition} We explore the impact of separately fine-tuning and evaluating both distilled and non-distilled efficient attention transformer models on CoNLL-2003 and GONERD in Table \ref{table:ner_results}. We report each model's F1 performance on predicting Person (PER), Organization (ORG), Location (LOC), and Miscellaneous (MISC) tags. We find that performing knowledge distillation prior to fine-tuning on NER preserved \textbf{97.4\%} of CoNLL-2003, while boosting GONERD F1 performance by \textbf{0.2\%}.

\subsection{Evaluating the Effect of Convert and Distill on Downstream Performance}
\input{tables/summary_distillation_convert}
To gauge the contribution of each component of the \textbf{Convert-Then-Distill} pipeline, we provide the computational cost and performance with respect to RoBERTa after undergoing conversion and distillation in Table \ref{table:summary_distillation_effects}. In contrast to Table \ref{table:ablation_extending}, the inference speeds and max GPU memory usages are calculated on sequences of up to 512 tokens. Within this range, we see that KD greatly improves inference speed while resulting in a minor decrease in maximum GPU memory utilization. Conversely, we see conversion to an efficient attention mechanism (in this case, Longformer) yields significant decreases in maximum GPU memory utilization and minor improvements in inference speed. Together, we find that \textit{Convert}+KD is additive in its effects: performing Conversion and KD yields models with both improved inference speeds and reduced GPU memory requirements,

We find long-context QA performance is heavily degraded by introducing \textit{Convert}+KD into training in comparison to other tasks, whereas conversion does not significantly impact performance. However, long-context NER appears to be an exception, as introducing \textit{Convert}+KD into GONERD has a significantly lower impact on performance. Finally, we note that the distillation process as used in DistilBERT \cite{sanh2019distilbert} and detailed in Appendix \ref{appendix:kd_details} leaves room for further improvement: developing distillation methods tailored for individual efficient attention mechanisms, tasks, and architectures may yield improved performance.

\subsection{Evaluating the Effect of Distillation Data on Downstream Performance}
\label{subsection:effect_of_distillation_data}
\input{tables/ablation_distillation_data}

For a more comprehensive evaluation of our knowledge distillation process, we report the performance of Longformer-RoBERTa after distillation on various permutations of the OSCAR, BookCorpus, and English Wikipedia datasets, as seen in Table \ref{table:ablation_data_kd}.

We find that, although OSCAR+BookCorpus yields the best performance on both short-context and long-context tasks, the performance gap between OSCAR+BookCorpus, Wikipedia+BookCorpus, and OSCAR+Wikipedia is very modest. However, as OSCAR+BookCorpus proved to be the most performant, we utilize this dataset when distilling efficient attention models.

%% file: tables/inference_memory.tex
\begin{table}[H]
\setlength{\belowcaptionskip}{-15pt}
\setlength{\tabcolsep}{3pt}
\begin{center}
{\footnotesize
\begin{tabular}{@{}cl@{}ccc@{\hskip .3cm}cc}
\cline{2-7}
& \multirow{2}{*}[-0.2ex]{Model} & \textbf{Params} & \multicolumn{2}{c}{\hspace{0cm}\textbf{Time (sec)}} & \multicolumn{2}{c}{\hspace{-.03cm}\textbf{Mem. (MB)}} \Tstrut\Bstrut\\
& & \textbf{(mil.)} & 512 & 4096 & 512 & 4096\Bstrut\\
\cline{2-7} \Tstrut
\multirow{5}{*}[-0.4ex]{\rotatebox{90}{\textbf{Baseline}}} & BERT$_{\textsc{BASE}}$ & \textbf{109.5} & \textbf{0.135} & - & \textbf{4167} & - \\
& BERT$_{\textsc{LARGE}}$ & 335.1 & 0.379 & - & 5171 & - \\
& RoBERTa & 124.6 & 0.148 & - & 4843 & - \\
& LegalBERT$_{\textsc{BASE}}$ & \textbf{109.5} & \textbf{0.135} & - & \textbf{4167} & - \\ \Bstrut
& XLM-R & 278.0 & 0.237 & - & 11673 & - \\
\cline{2-7} \Tstrut
\multirow{5}{*}[-0.3ex]{\rotatebox{90}{\textbf{Compressed}}} & DistilRoBERTa & 82.1 & 0.089 & - & 4663 & - \\
& DistilBERT & 66.4 & 0.078 & - & 3987 & - \\
& TinyBERT & \textbf{4.4} & \textbf{0.057} & - & \textbf{3033} & - \\
& MobileBERT & 24.6 & 0.072 & - & 3639 & - \\ \Bstrut
& ALBERT & 11.7 & 0.128 & - & 3783 & - \\
\cline{2-7} \Tstrut
\multirow{10}{*}[2.5ex]{\rotatebox{90}{\textbf{Efficient}}} & LSG & 127.8 & 0.170 & 1.157 & 5472 & 23482 \\
& \ding{229} & 85.3 & 0.103 & 0.641 & 5292 & 23302 \\
& Nyströmformer & 111.2 & 0.159 & 1.866 & 4291 & 29059 \\
& \ding{229} & \textbf{68.7} & 0.090 & 0.787 & 4111 & 28879 \\
& Longformer & 148.7 & 0.149 & 1.110 & 4077 & 11881 \\
& \ding{229} & 95.5 & \textbf{0.075} & \textbf{0.588} & \textbf{3857} & \textbf{11661} \\
& Big Bird & 127.5 & 0.158 & 1.542 & 4938 & 26854 \\ \Bstrut
& \ding{229} & 84.9 & 0.097 & 0.913 & 4757 & 26673 \\
\cline{2-7}
\end{tabular}
\vspace{.3cm}
\caption{Average Inference Speed and Peak GPU Memory Usage for sequence lengths of 512 and 4096. "\ding{229}" indicates distillation.}
\label{table:inference_memory}
}
\end{center}
\end{table}

%% file: tables/glue_performance.tex
\begin{table*}[ht]
\setlength{\belowcaptionskip}{-15pt}
\begin{center}
{\scriptsize
\begin{tabular}{lccccccccc}
\hline \Tstrut
\textbf{Model} & \textbf{CoLA} & \textbf{MNLI} & \textbf{MRPC} & \textbf{QNLI} & \textbf{QQP} & \textbf{RTE} & \textbf{SST-2} & \textbf{STS-B} & \textbf{Total}\\[0.5em]\Bstrut
Metric & MCC & M/MM Acc. & Acc./F1 & Acc. & Acc./F1 & Acc. & Acc. & PCC/SRCC & Avg. \\
\hline \Tstrut
BERT$_{\textsc{BASE}}^1$ & 52.1 & 84.6 / 83.4 & 88.9 & 90.5 & 71.2 &  66.4 & 93.5 & 85.8 & 79.6 \\
BERT$_{\textsc{LARGE}}^1$ & 60.5 & 86.7 / 85.9 & 89.3 & 92.7 & 72.1 & 70.1 & \textbf{94.9} & 86.5 & 82.1 \\
RoBERTa$^2$ & \textbf{63.6} & \textbf{87.6} & \textbf{90.2} & \textbf{92.8} & \textbf{91.9} & \textbf{78.7} & 94.8 & \textbf{91.2} & \textbf{86.4} \\
LegalBERT & 38.6 & 82.2 / 82.9 & 88.2 & 89.9 & 89.7 & 65.3 & 91.5 & 87.0 / 86.6 & 80.2 \\
\Bstrut
XLM-R & 59.8 & 85.3 / 85.7 & 88.2 / \textbf{91.6} & 92.3 & 90.7 / 87.6 & 77.3 & 93.3 & 90.9 / 90.6 & 86.1 \\[.5ex]
\hline \Tstrut 
DistilRoBERTa$^2$ & 59.3 & 84.0 & 86.6 & 90.8 & 89.4 & 67.9 & 92.5 & 88.3 & 82.4 \\
DistilBERT$^1$ & 52.4 & 82.6 & 86.5 & 89.5 & \textbf{88.6} & 60.3 & 91.3 & 86.8 & 79.8 \\
TinyBERT$^1$ & 43.3 & 82.5 / 81.8 & 86.4 & 87.7 & 71.3 & 62.9 & 92.6 & 79.9 & 76.5 \\
MobileBERT$_{\textsc{BASE}}^1$ & 50.5 & 83.3 / 82.6 & 88.8 & 90.6 & 70.2 & 66.2 & 92.8 & 84.4 & 78.8 \\
\Bstrut
ALBERT & \textbf{59.8} & \textbf{85.3} / \textbf{85.7} & \textbf{88.2 / 91.6} & \textbf{92.3} & \textbf{90.6} / 87.7 & \textbf{77.3} & \textbf{92.9} & \textbf{90.9 / 90.6} & \textbf{86.1} \\
\hline
%%%%%%%%%%%%%%%%%%%%%%%%%%%%%%%%%%%%%%%%%%%%%%%%% LSG %%%%%%%%%%%%%%%%%%%%%%%%%%%%%%%%%%%%%%%%%%%%%%%%
%%%%%%%%%%%%%%%%%%%% CoLA & MNLI & MRPC & QNLI & QQP & RTE & SST-2 & STS-B & WNLI %%%%%%%%%%%%%%%%%%%%
\Tstrut LSG & 59.8 & 86.7 / 86.1 & 89.7 / 92.5 & \textbf{93.4} & 89.8 / 86.3 & 70.0 & 94.8 & 90.2 / 90.0 & 84.2 \\
\ding{229} & 29.4 & 71.8 / 72.6 & 77.5 / 85.0 & 84.1 & 86.1 / 81.9 & 58.9 & 89.7 & 80.9 / 80.8 & 74.9 \\[.5em]
%%%%%%%%%%%%%%%%%%%%%%%%%%%%%%%%%%%%%%%%%%%% NYSTROMFORMER %%%%%%%%%%%%%%%%%%%%%%%%%%%%%%%%%%%%%%%%%%%
%%%%%%%%%%%%%%%%%%%% CoLA & MNLI & MRPC & QNLI & QQP & RTE & SST-2 & STS-B & WNLI %%%%%%%%%%%%%%%%%%%%
Nystr\"omformer & 33.6 & 77.9 / 79.1 & 77.7 / 84.7 & 86.3 & 88.8 / 85.0 & 56.7 & 90.8 & 86.1 / 86.0 & 77.7 \\
\ding{229} & 43.4 & 78.6 / 78.6 & 75.2 / 83.8 & 85.8 & 89.3 / 85.9 & 58.5 & 90.8 & 86.2 / 85.8 & 78.5 \\[.5em]
%%%%%%%%%%%%%%%%%%%%%%%%%%%%%%%%%%%%%%%%%%%%% LONGFORMER %%%%%%%%%%%%%%%%%%%%%%%%%%%%%%%%%%%%%%%%%%%%%
%%%%%%%%%%%%%%%%%%%% CoLA & MNLI & MRPC & QNLI & QQP & RTE & SST-2 & STS-B & WNLI %%%%%%%%%%%%%%%%%%%%
Longformer & \textbf{61.3} & 86.3 / 86.4 & \textbf{91.9} / \textbf{94.2} & 92.9 & 89.6 / 86.0 & \textbf{77.6} & 93.9 & \textbf{90.8} / \textbf{90.5} & \textbf{86.8} \\
\ding{229} & 55.5 & 82.0 / 81.8 & 82.1 / 86.9 & 87.7 & \textbf{90.3} / 86.8 & 54.2 & 91.7 & 86.2 / 86.0 & 80.9 \\[.5em]
%%%%%%%%%%%%%%%%%%%%%%%%%%%%%%%%%%%%%%%%%%%%%% BIGBIRD %%%%%%%%%%%%%%%%%%%%%%%%%%%%%%%%%%%%%%%%%%%%%%%
%%%%%%%%%%%%%%%%%%%% CoLA & MNLI & MRPC & QNLI & QQP & RTE & SST-2 & STS-B & WNLI %%%%%%%%%%%%%%%%%%%%
Big Bird & 51.6 & \textbf{87.1} / \textbf{87.3} & 87.8 / 91.3 & 91.0 & \textbf{90.3} / \textbf{86.9} & 68.2 & \textbf{95.0} & 86.5 / 86.5 & 84.1 \\
\Bstrut \ding{229} & 53.9 & 81.6 / 81.9 & 82.4 / 87.3 & 86.8 & 90.2 / 86.5 & 59.6 & 92.4 & 85.2 / 84.8 & 81.0 \\
\hline
\end{tabular}
\vspace{.3cm}
\caption{Results on the validation set of the GLUE benchmark. "\ding{229}" indicates distillation performance. Results for $^{1,2}$ are pulled from MobileBERT \cite{sun2020mobilebert} and  DistilBERT \cite{sanh2019distilbert} papers, respectively; all other models are computed to completion. WNLI is not reported due to its problematic nature \cite{devlin2018bert}.}
\label{table:glue_results}

}
\end{center}
\end{table*}

%% file: tables/extend_input_performance.tex
\begin{table}[h!]
\setlength{\belowcaptionskip}{-10pt}
\setlength{\tabcolsep}{9pt}
\begin{center}
{\scriptsize
\begin{tabular}{lcc}
\hline \Tstrut
Model$^{\uparrow4096}$ & \textbf{Time (sec)} & \textbf{GPU Mem (MB)} \\[.4ex]
\hline \Tstrut
BERT$_{\textsc{LARGE}}$ & 5.806 (+423\%) & 39344 (+221\%) \\
BERT$_{\textsc{BASE}}$ & 1.833 (+65\%) & 29886 (+152\%) \\
RoBERTa & 1.886 (+70\%) & 42506 (+258\%) \\
DistilBERT & 0.636 (+8\%) & 29706 (+155\%) \\
DistilRoBERTa & 0.798 (+36\%) & 42326 (+163\%) \\
MobileBERT & 1.274 (+117\%) & \textbf{24406 (+109\%)} \\\Bstrut
TinyBERT & \textbf{0.631 (+7\%)} & 29706 (+155\%) \\ 
\hline \Tstrut
Longformer & 1.110 & 11881 \\
\ding{229} & \textbf{0.588} & \textbf{11661} \\
\hline
\end{tabular}
\vspace{.3cm}
\caption{Inference speed and GPU memory consumption when extending the maximum input sequence length from 512 to 4096 for various models. Percentages for non-compressed models are calculated against Longformer, while percentages for compressed models are calculated against distilled Longformer.}
\label{table:ablation_extending}
}
\end{center}
\end{table}

%% file: tables/qa_performance.tex
\begin{table}[h!]
\setlength{\belowcaptionskip}{-20pt}
\setlength{\tabcolsep}{4pt}
\begin{center}
{\scriptsize
\begin{tabular}{l@{\hskip .5cm}cc@{\hskip .5cm}cc@{\hskip .5cm}cc}
\hline \Tstrut
\multirow{2}{*}{Model} & \multicolumn{2}{c}{\hspace{-.5cm}\textbf{SQuAD1.1}} & \multicolumn{2}{c}{\hspace{-.4cm}\textbf{HotpotQA}} & \multicolumn{2}{c}{\hspace{-.2cm}\textbf{TriviaQA}} \\ \Bstrut
                                  & EM & F1 & EM & F1 & EM & F1 \\ \hline \Tstrut
BERT$_{\textsc{BASE}}$ & 80.97 & 88.21 & - & - & - & - \\
BERT$_{\textsc{LARGE}}$ & 83.91 & 90.73 & - & - & - & - \\
RoBERTa & \textbf{86.08} & \textbf{92.47} & - & - & - & - \\
LegalBERT & 79.89 & 87.66 & - & - & - & - \\ \Bstrut
XLM-R & 82.38 & 89.16 & - & - & - & - \\
\hline \Tstrut
DistilRoBERTa & 80.43 & 87.87 & - & - & - & - \\
DistilBERT & 77.01 & 85.21 & - & - & - & - \\
TinyBERT & 69.77 & 78.89 & - & - & - & - \\
MobileBERT & 80.83 & 88.56 & - & - & - & - \\ \Bstrut
ALBERT & \textbf{83.58} & \textbf{90.64} & - & - & - & - \\
\hline \Tstrut
LSG & 80.61 & 87.89 & 56.96 & 72.11 & 47.34 & 51.82 \\
\ding{229} & 64.20 & 74.13 & 41.77 & 54.58 & 26.67 & 30.00 \\
Nyströmformer & 76.59 & 84.89 & 52.57 & 67.86 & 47.30 & 52.29 \\
\ding{229} & 70.87 & 80.51 & 48.54 & 63.87 & 44.55 & 49.68 \\
Longformer & \textbf{85.92} & \textbf{92.24} & 58.52 & 73.48 & \textbf{55.29} & \textbf{60.52} \\ 
\ding{229} & 77.93 & 85.81 & 49.86 & 64.96 & 46.75 & 51.42 \\
Big Bird & 84.94 & 91.44 & \textbf{59.77} & \textbf{75.26} & 54.29 & 59.33 \\ \Bstrut 
\ding{229} & 74.53 & 82.67 & 49.40 & 64.21 & 44.61 & 49.96 \\
\hline
\end{tabular}
\vspace{.3cm}
\caption{SQuAD, HotpotQA, and TriviaQA Results.}
\label{table:qa_performance}
}
\end{center}
\end{table}

%% file: tables/ner_performance.tex
\begin{table}[ht!]
\setlength{\belowcaptionskip}{-20pt}
\setlength{\tabcolsep}{2pt}
\begin{center}
{\scriptsize
\begin{tabular}{lccccc@{\hskip 8pt}ccccc}
\hline
\multirow{2}{*}{\raisebox{-3.8ex}{Model}} & \multicolumn{5}{c}{\textbf{CoNLL-2003} (512)} & \multicolumn{5}{c}{\textbf{GONERD} (4096)}\Tstrut\\
 & \rotatebox{65}{\textbf{PER}} & \rotatebox{65}{\textbf{ORG}} & \rotatebox{65}{\textbf{LOC}} & \rotatebox{65}{\textbf{MISC}} & \rotatebox{65}{\textbf{ALL}} & \rotatebox{65}{\textbf{PER}} & \rotatebox{65}{\textbf{ORG}} & \rotatebox{65}{\textbf{LOC}} & \rotatebox{65}{\textbf{MISC}} & \rotatebox{65}{\textbf{ALL}}\Bstrut\\
\hline \Tstrut
BERT$_{\textsc{BASE}}$ & 97.1 & 89.8 & 95.4 & 87.9 & 92.6 & - & - & - & - & - \\
BERT$_{\textsc{LARGE}}$ & \textbf{98.6} & \textbf{92.6} & \textbf{96.5} & \textbf{88.8} & \textbf{94.1} & - & - & - & - & - \\
RoBERTa & 96.2 & 91.3 & 96.4 & 89.8 & 93.4 & - & - & - & - & - \\
LegalBERT & 95.3 & 87.2 & 94.8 & 86.0 & 90.8 & - & - & - & - & - \\ \Bstrut
XLM-R & 95.6 & 90.2 & 95.8 & 89.8 & 92.9 & - & - & - & - & - \\
\hline \Tstrut
DistilRoBERTA & \textbf{96.7} & \textbf{92.1} & \textbf{96.7} & \textbf{90.1} & \textbf{93.9} & - & - & - & - & -\\
DistilBERT & 96.7 & 89.2 & 95.4 & 88.3 & 92.4 & - & - & - & - & - \\
TinyBERT & 95.6 & 87.8 & 95.3 & 86.8 & 91.4 & - & - & - & - & - \\
MobileBERT & 97.8 & 90.2 & 96.4 & 87.9 & 93.1 & - & - & - & - & - \\ \Bstrut
ALBERT & 93.8 & 85.6 & 94.5 & 86.3 & 90.1 & - & - & - & - & - \\
\hline \Tstrut
LSG & 96.6 & 90.0 & 95.2 & 88.1 & 92.5 & \textbf{76.5} & 66.7 & 64.0 & \textbf{78.7} & 70.2 \\
\ding{229} & 89.8 & 80.0 & 92.2 & 81.3 & 85.8 & 69.8 & 59.0 & 60.7 & 72.6 & 64.1 \\
Nyströmformer & 94.8 & 85.1 & 93.3 & 86.4 & 89.9 & 72.4 & 59.5 & 59.6 & 75.1 & 65.0 \\
\ding{229} & 95.3 & 85.1 & 94.2 & 85.6 & 90.1 & 70.6 & 56.4 & 60.2 & 70.5 & 63.3  \\
Longformer & 96.2 & \textbf{91.5} & \textbf{96.8} & \textbf{90.5} & \textbf{93.8} & 75.9 & \textbf{68.0} & 65.1 & 77.3 & \textbf{70.6} \\
\ding{229} & \textbf{96.7} & 91.2 & 96.7 & 89.8 & 93.6 & 71.8 & 65.1 & 63.3 & 76.3 & 67.7 \\
Big Bird & 96.4 & 91.8 & 96.4 & 89.8 & 93.6 & 75.9 & 65.4 & \textbf{66.3} & 73.1 & 69.8 \\ \Bstrut
\ding{229} & 96.2 & 90.4 & 96.2 & 89.7 & 93.1 & 71.6 & 63.2 & 61.7 & 73.2 & 66.4 \\
\hline
\end{tabular}
\vspace{.3cm}
\caption{\textbf{N}amed \textbf{E}ntity \textbf{R}ecognition (\textbf{NER}) F1 Performance on CoNLL-2003 and GONERD.}
\label{table:ner_results}
}
\end{center}
\end{table}

%% file: tables/summary_distillation_convert.tex
\begin{table}[H]
\setlength{\belowcaptionskip}{-10pt}
\setlength{\tabcolsep}{.7pt}
\begin{center}
{\scriptsize
\begin{tabular}{l@{\hskip .13cm}ccccccccc}
\hline 
 \\[-.2cm]
\raisebox{.5cm}{Model} & \rotatebox{70}{\textbf{Inf. Time (sec)}} & \rotatebox{70}{\textbf{GPU Mem.}} & \rotatebox{70}{\textbf{GLUE}} & \rotatebox{70}{\textbf{SQuAD1.1}} & \rotatebox{70}{\textbf{CoNLL-2003}} & \rotatebox{70}{\textbf{HotpotQA}} & \rotatebox{70}{\textbf{TriviaQA}} & \rotatebox{70}{\textbf{GONERD}} \Tstrut\Bstrut\\
\hline \Tstrut
RoBERTa & .148 & 4843 & 86.35 & 92.47 & 93.43 & - & - & - \\
$\Delta$ KD & -39.9\% & -3.7\% & -4.6\% & -5.0\% & +0.5\% & - & - & - \\
$\Delta$ \textit{Convert} & +0.7\% & -15.8\% & -0.5\% & -2.5\% & +0.3\% & 73.48 & 60.52 & 70.6 \\ \Bstrut
$\Delta$ \textit{Convert}+KD & \textbf{-49.3\%} & \textbf{-20.4\%} & \textbf{-6.3\%} & \textbf{-7.0\%} &  \textbf{+0.2\%} & \textbf{-11.6\%} & \textbf{-15.0\%} & \textbf{-4.1\%} \\
\hline
\end{tabular}
\vspace{.3cm}
\caption{Effects of introducing Knowledge Distillation and Longformer attention into RoBERTa on various tasks. We report average score for GLUE and overall F1 for QA and NER. "$\Delta$" indicates a change from the base model. Results are compared against RoBERTa on short-context tasks and against Longformer ($\Delta$ \textit{Convert}) on long-context tasks. Sequence lengths of 512 are used for inference time and memory usage.}
\label{table:summary_distillation_effects}
}
\end{center}
\end{table}

%% file: tables/ablation_distillation_data.tex
\begin{table}[ht!]
\setlength{\belowcaptionskip}{-10pt}
\setlength{\tabcolsep}{1pt}
\begin{center}
{\scriptsize
\begin{tabular}{l@{\hskip .4cm}ccccc}
\hline 
 \\[-2ex]
\raisebox{.5cm}{Distillation Data} & \rotatebox{65}{\textbf{GLUE}} & \rotatebox{65}{\textbf{SQuAD1.1}} & \rotatebox{65}{\textbf{HotpotQA}} & \rotatebox{65}{\textbf{CoNLL-2003}} & \rotatebox{65}{\textbf{GONERD}} \\
\hline
\textsc{BC + WIKI} & 75.6 & 77.9 & 57.7 & 92.2 & 65.5 \Tstrut\\
\textbf{\textsc{OSCAR + BC}} & \textbf{78.9} & \textbf{85.8} & \textbf{65.0} & \textbf{93.6} & \textbf{67.7} \\
\textsc{OSCAR + WIKI} & 77.1 & 78.8 & 60.6 & 92.8 & 66.3 \\
\textsc{WIKI} & 68.7 & 76.5 & 57.9 & 91.9 & 63.8 \\
\textsc{OSCAR} & 60.5 & 25.4 & 39.6 & 73.5 & 38.3 \\
\textsc{BC} & 72.9 & 56.4 & 40.7 & 90.4 & 40.0 \Bstrut\\
\hline
\end{tabular}
\vspace{.3cm}
\caption{Effects of choice of data on KD performance using Longformer RoBERTa with a train batch size of 4. Average score, not including WNLI, is reported for GLUE and overall F1 is reported for QA and NER. A full expansion of GLUE results is given in Table \ref{table:ablation_glue_results}.}
\label{table:ablation_data_kd}
}
\end{center}
\end{table}

%% file: sections/5_conclusion.tex
\section{Conclusion}

In this work, we performed an investigation into the \textbf{Convert-Then-Distill} paradigm, the process of (1) \textit{converting} a teacher model to utilize an efficient attention mechanism, (2) pretraining the converted teacher model, (3) distilling into a smaller student model, then (4) fine-tuning the student on a downstream task. We saw an average decrease in inference times of up to \textbf{58\%}. The efficient attention students preserved up to \textbf{98.6\%} of performance across short-context (GLUE, SQuAD, CoNLL-2003) tasks, \textbf{94.1\%} of HotpotQA performance, \textbf{95.0\%} of TriviaQA performance, and \textbf{97.4\%} percent of GONERD performance when compared to their teacher models. We saw distilled  Nyströmformer retained the most performance when compared to its teacher, while distilled Longformer had the best base performance across most tasks. We introduced GONERD, a long-context NER dataset consisting of large amounts of hand-labeled web text data. Finally, we release all models on the Hugging Face Hub for general use. Our research demonstrates that, for most models on most tasks, employing knowledge distillation on efficient attention architectures can be a highly effective approach. This technique yields models with a high level of performance on both short and long-context tasks at a fraction of the cost.

%% file: sections/7_acknowledgements.tex
\section*{Acknowledgements}
This work was produced through a partnership between Clemson University and Giant Oak. We thank Gary Shiffman and Carrie Russell for their invaluable mentorship and support. We are indebted to the data assembly efforts and guidance of the Giant Oak research assistants and science members, including but not limited to Marena Dangremond, Oladotun Taiwo, Omar Ocasio, Rohan Jani, Tyler Strickland, Kateri Gajadhar-Smith, Alexa Wingate, Timothy Ressler, and Benjamin Crisman. Clemson University is acknowledged for their generous allotment of compute time on the Palmetto Cluster. We appreciate D. Hudson Smith for his assistance and comments as well as the insightful comments of the anonymous reviewers. This material is based on work supported by the National Science Foundation under Grant Nos. MRI\# 2024205, MRI\# 1725573, and CRI\# 2010270. Any opinions, findings, and conclusions or recommendations expressed in this material are those of the author(s) and do not necessarily reflect the views of the National Science Foundation.\\

%% file: sections/8_limitations.tex
\section*{Limitations}

 As seen in Section \ref{subsection:performance_benchmarks}, we find that, in both short and long sequences, \textbf{Convert-Then-Distill} degrades performance to a greater extent than either Convert or KD separately. This performance degradation warrants further investigation into generalization capabilities of efficient students.

Following this, many distillation procedures have been proposed since the original technique of DistilBERT \cite{sanh2019distilbert}. Using more recent distillation methods, or developing distillation methods tailored toward an individual efficient attention architecture, may decrease the student-teacher performance gap and increase generalizability.

Our work is constrained to the \textbf{Convert-Then-Distill} paradigm which, although intuitive, is not obviously better than \textbf{Distill-Then-Convert} or other alternatives. For instance, it may be possible that non-efficient teachers produce better students which can then be extended to the 4096 or greater token range. Further investigation into the optimal method for developing distilled efficient attention models may be necessary to further close the aforementioned performance gap.

Finally, GONERD suffers from a domain bias as it is composed entirely of news-like webtext data and commonly littered with legal jargon. We attempt to control for this bias by comparing with LegalBERT and ablating on choices of pretraining data, but we note this bias for any potential users of GONERD. For general long-context NER use, additional pretraining may be required.

%% file: sections/6_appendix.tex
\section{GONERD}
\label{appendix:gonerd}

\subsection{Data Collection}
\label{appendix:gonerd_data_collection}

Data for GONERD was obtained through Giant Oak's GONER software, which scraped web articles from public facing online news sources as well as the U.S. Department of Justice's \textit{justice.gov} domain. This webtext data was randomly sampled with an upweighted probability toward documents from \textit{justice.gov} so that \textit{justice.gov} consisted of roughly 25\% of the total GONERD dataset. A combination of automatic and manual filtering was then applied to remove text containing code and other unwanted data, such as sequences deemed to be short, noisy, or near-duplicates. 

\subsection{Exploratory Data Analysis}
\label{appendix:gonerd_eda}
\input{tables/gonerd_conll_types}
\input{figures/length_gonerd_conll}

\paragraph{Sequence Length Distribution} In Figure \ref{fig:gonerd_length_dist}, we display the distribution of CoNLL-2003 sequences in orange and GONERD sequences in blue. To produce the figure, we used standard Kernel Density Estimation (KDE) through the \textit{kdeplot} function of the Python seaborn library. For the GONERD distribution, we used the default parameters of the \textit{kdeplot} function, but for CoNLL-2003, we used a higher KDE bandwidth and upsampled the distribution in the 256 range, thereby giving CoNLL-2003 a slightly synthetically higher distribution, resulting in CoNLL-2003 sequences appearing to be longer than they actually are. We perform this to account for to the extreme gap in average sequence length between CoNLL-2003 and GONERD. CoNLL-2003 has a large number of short sequences which make the table significantly taller, making visually comparing their distributions unintelligible. We briefly provide summary statistics in Table \ref{table:length_stats_conll_gonerd} to evidence this gap. 

\input{tables/gonerd_domains}

\paragraph{Entity Makeup} For our NER task, we evaluated the distribution of tags to gain a deeper understanding of our evaluation results. As seen in Tables \ref{table:gonerd_entity_makeup} and \ref{table:gonerd_domains}, although ConLL-2003 consists of more sequences, GONERD has $3.5\times$ as many labeled tokens. Additionally, we find that names in GONERD tend to be longer than CoNLL-2003, as evidenced by the proportion of I to B tokens across all NER tags. For GONERD, we find this proportion to be 57.3/64.7 in comparison to 15.6/35 for CoNLL-2003.

As seen in Table \ref{table:ner_results}, LOC and ORG are the most difficult for both teacher and student teachers to learn in GONERD. This may come as a surprise when considering the MISC tag, in which all efficient attention models obtained better performance despite MISC's fewer samples. One possible explanation for the discrepancy in MISC performance is in how GONERD handles MISC labeling. GONERD has a fixed schema for MISC: ages, addresses, and phone numbers, while everything else is not marked as a valid entity. As this reduces the diversity of this category, this could make the MISC tag easier for models to learn to detect. This is in stark comparison to CoNLL-2003, in which MISC consists of adjectives and events, making it a very diverse category. This can be evidenced by the performance difference for efficient attention models on CoNLL-2003, where MISC was the most difficult tag for models to learn. Outside of the discrepancy on MISC, GONERD's makeup closely resembles ConLL-2003 regarding the distribution of non-O tags but leverages long-context, making it a valuable asset to long-context NER models.

\paragraph{Domains} In Table \ref{table:gonerd_domains}, we show the frequencies of the top ten domains occurring in GONERD, ranked by relative occurrence. The raw number of samples under a domain is denoted by "\#", the relative proportion by "pdf," and the cumulative by "cdf."

Aligning with expectations, we see that \textit{justice.gov} appears in \textbf{24.2\%} samples, a website full of news and legal language, primarily in the form of criminal charges and sentencing. However, as domains progress, the relative contribution drops off exponentially, with the top ten domains only making up \textbf{48.9\%} of GONERD whereas there are 369 domains present within the dataset. \\

\section{Knowledge Distillation Details}
\label{appendix:kd_details}

Briefly, we give an overview of the student objective used in our distillation experiments, which we frame as the linear combination of \emph{supervised training} loss, \emph{distillation} loss, and \emph{hidden state} loss. Our supervised training loss is the standard masked language modeling loss \cite{devlin2018bert}. Our distillation loss is a cross entropy over soft targets \cite{hinton2015distilling, sanh2019distilbert}, which are calculated by applying a softmax with temperature to the output logits: 
\begin{align*}
    p_i = \frac{exp(z_i/T)}{\sum_{j}exp(z_j/T)}
\end{align*}
where $p_i$ is the probability of logit $z_i$ and $T$ is temperature, which controls the smoothness of the distribution. Following the methods outlined in DistilBERT \cite{sanh2019distilbert}, we use a cosine embedding loss between the hidden states vectors of the teacher and student as a hidden state loss. Our overall training objective can thus be written as
\begin{align*}
        \mathcal{L}_{student} &= \alpha\mathcal{L}_{mlm} + \beta\mathcal{L}_{ce} + \gamma\mathcal{L}_{cse}
\end{align*}
We take $\alpha = 2.0$, $\beta = 5.0$, $\gamma = 1.0$, and $T = 2.0$. Finally, we train the student by minimizing the associated empirical risk with the AdamW optimizer.

\section{Data Ablation Results}
\label{appendix:data_ablation_glue_full}

Finally, we expand upon the GLUE performance given in Table \ref{table:ablation_data_kd}, distilling Longformer RoBERTa on various permutations of the BookCorpus (BC), English Wikipedia (ENW), and OSCAR datasets and evaluating on all GLUE tasks. All models are trained identically as given in Section \ref{subsection:effect_of_distillation_data}.

We find that distilling Longformer on OSCAR and BookCorpus yields the highest GLUE scores, with an average of \textbf{78.9} across all tasks and metrics. However, both BookCorpus and English Wikipedia as well as OSCAR and English Wikipedia still yield very similar results, with the most notable differences being in the CoLA and MNLI tasks. We see significantly lower scores, particularly on CoLA, when Longformer is distilled using only short or long sequences. This indicates that it may be necessary for efficient attention models to be distilled using a mixture of both short and long-context data to ensure maximum student performance.
\input{tables/ablation_distillation_glue}

%% file: tables/gonerd_conll_types.tex
\begin{table}[h!]
\setlength{\belowcaptionskip}{-20pt}
\begin{center}
{\small
\begin{tabular}{l@{\hskip .5cm}c@{\hskip .2cm}c@{\hskip .2cm}c@{\hskip .6cm}c@{\hskip .2cm}c@{\hskip .2cm}c@{\hskip .1cm}}
    \hline
    \multirow{2}{*}{Type} & \multicolumn{3}{c}{\hspace{-0.5cm}\textbf{CoNLL-2003} (512)} & \multicolumn{3}{c}{\hspace{-.2cm}\textbf{GONERD} (4096)} \Tstrut\\
     & \# & $p$ & $p^1$ & \# & $p$ & $p^1$ \Bstrut\\
    \hline
    O & 251k & .832 & - & 1013k & .896 & - \Tstrut\\
    B-PER & 10k & .033 & .198 & 24.4k & .022 & .200 \\
    I-PER & 6.9k & .023 & .138 & 22.9k & .020 & .188 \\
    B-ORG & 9.3k & .031 & .184 & 21.5k & .019 & .177 \\
    I-ORG & 5.3k & .018 & .104 & 23.0k & .020 & .188 \\
    B-LOC & 10.6k & .035 & .210 & 16.4k & .014 & .134 \\
    I-LOC & 1.7k & .006 & .033 & 9.7k & .009 & .079 \\
    B-MISC & 5.1k & .017 & .100 & 2.4k & .002 & .020 \\ \Bstrut
    I-MISC & 1.7k & .006 & .034 & 1.7k & .002 & .014 \\
    \hline \Tstrut \Bstrut
    \textbf{Total} & 301k & 1.0 & 1.0 & 1131k & 1.0 & 1.0  \\
    \hline
    \end{tabular}
    \vspace{.3cm}
\caption{Occurrence of PER/ORG/LOC/MISC/O tags in CoNLL-2003 and GONERD. $p$ represents the proportion of a tag over the total amount of labeled tokens and $p^1$ represents the proportion over non-O tokens.}
\label{table:gonerd_entity_makeup}
}
\end{center}
\end{table}

%% file: figures/length_gonerd_conll.tex
\begin{figure*}[ht!]
	\centering
    \includegraphics[scale=.32]{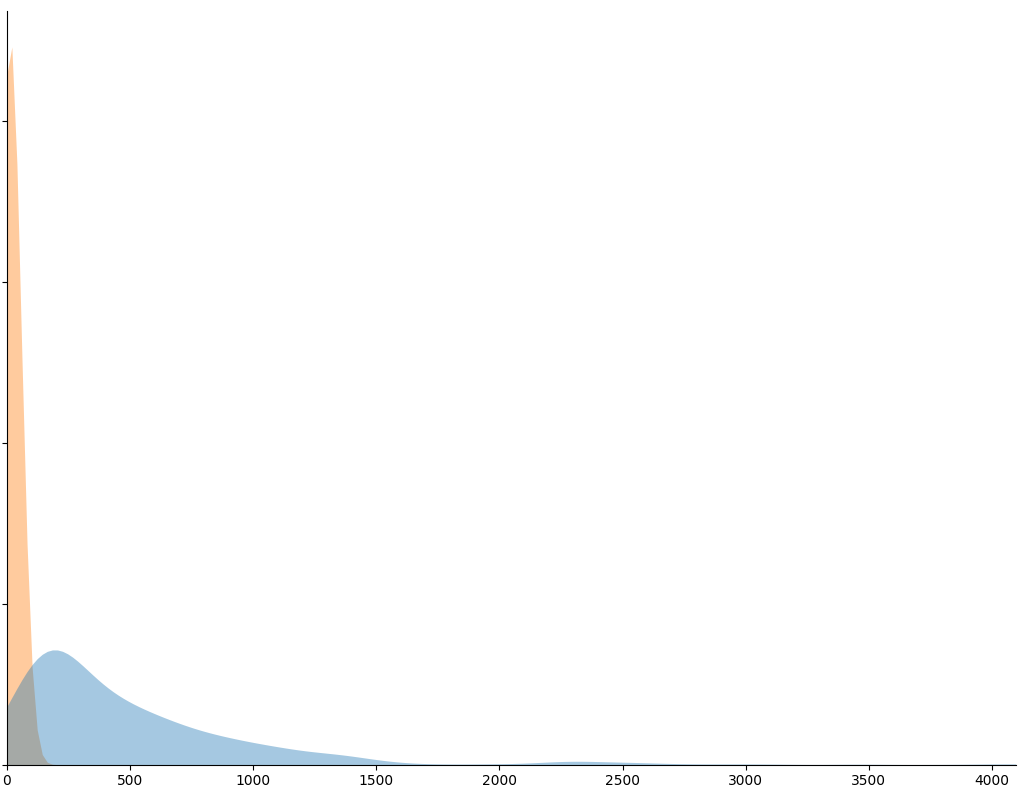}
	\caption{Sequence length distribution using Kernel Density Estimation (KDE) of CoNLL-2003 (orange) and GONERD (blue). Smoothing was performed with Gaussian KDE using the seaborn kdeplot function. x-axis is number of tokens and y-axis is probability density. }
	\label{fig:gonerd_length_dist}
\end{figure*}

%% file: tables/gonerd_domains.tex
\begin{table}[h!]
\setlength{\belowcaptionskip}{-15pt}
\setlength{\tabcolsep}{1pt}
\begin{center}
{\normalsize
\begin{tabular}{l@{\hskip .3cm}c@{\hskip .2cm}c@{\hskip .2cm}c}
\hline \Tstrut
Website & \# & pdf & cdf \\[.4ex]
\hline \Tstrut
justice.gov & 542 & .242 & .242 \\
ctvnews.ca & 190 & .085 & .327 \\
msn.com & 146 & .065 & .392 \\
southcarolinapublicradio.org & 54 & .024 & .416 \\
express.co.uk & 41 & .018 & .434 \\
dailyrecord.com & 33 & .015 & .449 \\
dailyvoice.com & 28 & .013 & .462 \\
nbcnews.com & 23 & .010 & .472 \\
newsbreak.com & 21 & .009 & .481 \\
chicagotribune.com & 19 & .008 & .489 \\[-1ex]
... & ... & ... & ... \\
\hline \Tstrut \Bstrut
\textbf{Total} & 2237 & 1.0 & 1.0 \\
\hline\end{tabular}
\vspace{.3cm}
\caption{Occurrence of domains in GONERD. "\#" is the raw amount of samples occuring under a domain, "pdf" is the proportion of samples in the whole dataset for that domain, and "cdf" is the cumulative proportion of samples sorted by frequency. Results are sorted by descending "pdf." Asteriscs indicate data not shown.}
\label{table:gonerd_domains}
}
\end{center}
\end{table}

%% file: tables/ablation_distillation_glue.tex
\onecolumn
\begin{table}[h]
\begin{center}
{\scriptsize
\begin{tabular}{lccccccccc}
\hline \Tstrut
\textbf{Model} & \textbf{CoLA} & \textbf{MNLI} & \textbf{MRPC} & \textbf{QNLI} & \textbf{QQP} & \textbf{RTE} & \textbf{SST-2} & \textbf{STS-B} & \textbf{Total}\\[0.5em]\Bstrut
Metric & MCC & M/MM Acc. & Acc./F1 & Acc. & Acc./F1 & Acc. & Acc. & PCC/SRCC & Avg. \\
\hline \Tstrut
\textsc{BC + ENW} & 41.7 & 77.3 / 77.6 & 82.6 / 87.6 & 83.9 & 88.0 / 84.7 & 56.7 & 89.7 & 84.4 / 84.1 & 75.6 \\
\textsc{OSCAR + BC} & \textbf{52.1} & \textbf{81.8} / \textbf{82.3} & \textbf{84.8} / \textbf{88.9} & \textbf{87.3} & \textbf{89.9} / \textbf{86.6} & 57.0 & \textbf{91.7} & \textbf{86.3} / \textbf{86.1} & \textbf{78.9} \\
\textsc{OSCAR + ENW} & 46.1 & 76.8 / 78.9 & 83.8 / \textbf{88.9} & 86.2 & 87.9 / 83.1 & \textbf{58.5} & 91.3 & 85.2 / 84.9 & 77.1 \\
\textsc{ENW} & 7.1 & 73.8 / 74.3 & 79.4 / 86.0 & 81.6 & 86.0 / 80.9 & 54.2 & 85.1 & 81.6 / 81.3 & 68.7 \\
\textsc{OSCAR} & 10.6 & 67.7 / 44.0 & 72.3 / 82.0 & 75.3 & 82.6 / 77.5 & 47.3 & 81.5 & 56.0 / 56.8 & 60.5 \\
\Bstrut \textsc{BC} & 38.7 & 74.2 / 75.6 & 75.7 / 83.6 & 83.1 & 87.1 / 82.5 & 55.2 & 88.3 & 78.6 / 78.3 & 72.9  \\
\hline
\end{tabular}
\vspace{.3cm}
\caption{Full validation results for GLUE on the students in the distillation ablative experiment in Section \ref{subsection:effect_of_distillation_data}.}
\label{table:ablation_glue_results}

}
\end{center}
\end{table}
\twocolumn